\pdfoutput=1

\documentclass[11pt]{article}

\usepackage[final]{acl}

\usepackage{times}
\usepackage{latexsym}

\usepackage[T1]{fontenc}

\usepackage[utf8]{inputenc}
\usepackage{url}
\usepackage{microtype}
\usepackage{booktabs}
\usepackage{times}
\usepackage{latexsym}
\usepackage{booktabs}
\usepackage{amsmath}
\usepackage{multirow}
\usepackage{tabularx}
\usepackage{caption}
\usepackage{rotating}
\usepackage{graphicx}
\usepackage{subfig}
\usepackage{tabularx}
\usepackage{adjustbox}
\usepackage{pgfplotstable}
\usepackage{xcolor,colortbl}
\usepackage{inconsolata}
\usepackage{makecell}

\usepackage{dblfloatfix}

\definecolor{myred}{RGB}{248, 141, 125}

\title{Unveiling the Generalization Power of Fine-Tuned Large Language Models}

\author{Haoran Yang$^{\spadesuit}$\hspace{1.0mm}, 
    Yumeng Zhang$^{\heartsuit}$\hspace{0.5mm},
    Jiaqi Xu$^{\spadesuit}$\hspace{0.5mm}, 
    \textbf{Hongyuan Lu}$^{\spadesuit}$\hspace{0.5mm}, \\ 
    \textbf{Pheng Ann Heng}$^{\spadesuit}$\hspace{0.5mm}, 
    \textbf{Wai Lam}$^{\spadesuit}$\hspace{0.5mm} 
    \hspace{0.5mm} \\
    $^{\spadesuit}$The Chinese University of Hong Kong~$^{\heartsuit}$Tsinghua University\\
    \texttt{\{hryang, hylu, wlam\}@se.cuhk.edu.hk \{jqxu,pheng\}@cse.cuhk.edu.hk} \\
    \texttt{zhang-ym23@mails.tsinghua.edu.cn} \\}

\begin{document}
\maketitle
\begin{abstract}
While Large Language Models (LLMs) have demonstrated exceptional multitasking abilities, fine-tuning these models on downstream, domain-specific datasets is often necessary to yield superior performance on test sets compared to their counterparts without fine-tuning. However, the comprehensive effects of fine-tuning on the LLMs' generalization ability are not fully understood.
This paper delves into the differences between original, unmodified LLMs and their fine-tuned variants. Our primary investigation centers on whether fine-tuning affects the generalization ability intrinsic to LLMs. To elaborate on this, we conduct extensive experiments across five distinct language tasks on various datasets.
Our main findings reveal that models fine-tuned on generation and classification tasks exhibit dissimilar behaviors in generalizing to different domains and tasks.
Intriguingly, we observe that integrating the in-context learning strategy during fine-tuning on generation tasks can enhance the model's generalization ability.
Through this systematic investigation, we aim to contribute valuable insights into the evolving landscape of fine-tuning practices for LLMs. The code and data are available at \url{https://github.com/LHRYANG/Generalization_of_FT-LLM}.

\end{abstract}

\section{Introduction}

The transformative impact of in-context learning (ICL)~\cite{NEURIPS2020_1457c0d6,wei2022chain,rubin-etal-2022-learning,pretrainprompt,chowdhery2022palm, wang2023hint} in LLMs, as demonstrated by models like Llama-2~\cite{touvron2023llama} and GPT-3~\cite{NEURIPS2020_1457c0d6}, marks a significant advancement in the field of artificial intelligence. 
This learning paradigm allows LLMs to adapt to various tasks by leveraging multiple demonstration examples presented within a prompt, without training the LLMs.
However, when it comes to a specific task, fine-tuning often achieves better performance than ICL, which has been substantiated by recent studies~\cite{shi2023specialist,jiao2023chatgpt,wang2023instructuie,zhang2023multitask}.

There are some works studying the properties of fine-tuning and ICL for LLMs.
For example, \citet{wei2022finetuned} reveal that \textit{multi-task} fine-tuning can enhance an LLM's zero-shot and ICL capabilities. It indicates that fine-tuning, when applied across multiple tasks, does not merely improve performance on those seen tasks but also augments the model's inherent learning abilities.
The work of \citet{mosbach-etal-2023-shot} highlights that in terms of the out-of-domain generalization in classification tasks, \textit{few-shot} fine-tuning and ICL exhibit similar levels of generalization. 
\citet{wang2023twostage} find that fine-tuning may overly tailor the model to task-specific formats, potentially compromising its adaptability to other new tasks.

In this paper, we conduct a comprehensive study on how \textit{task-specific} (not multi-task or few-shot) fine-tuning affects the generalization ability of LLMs.
To provide a thorough analysis, we design a series of experiments encompassing a diverse range of datasets and tasks, covering both classification and generation tasks.
For each task, we designate a specific dataset as the training set. The remaining selected datasets are subsequently divided into two groups: in-domain datasets, closely aligned with the training set in terms of content and structure, and out-of-domain datasets, which possess significant differences.
With these datasets, our research investigates two critical questions: i)~the ability of fine-tuned LLMs to adapt to both in-domain and out-of-domain test sets, and ii)~the impact of fine-tuning on the ICL ability of LLMs across different types of tasks.

We find that models fine-tuned on text generation and classification tasks exhibit different behaviors when evaluated on test sets.
Specifically, we observe that models fine-tuned for classification tasks tend to exhibit positive transfer when applied to out-of-domain datasets of the same fine-tuning/test task type. In contrast, models fine-tuned on generation tasks frequently experience negative transfer under similar conditions.
Interestingly, while fine-tuning the LLMs on generation tasks generally does not detrimentally affect their performance on classification tasks, the reverse is not true; models fine-tuned on classification tasks typically fail to work on generation tasks.
Moreover, we experimentally observe that integrating the ICL strategy during fine-tuning on generation tasks can enhance an LLM’s generalization ability. We also investigate other factors, such as training data size and the number of in-context examples.
We hope this study offers comprehensive insights into fine-tuning strategies for LLMs, not only in enhancing task-specific performance but also in fostering broader generalization abilities.


\section{Related Work}

\subsection{Large Language Models}
The emergence and evolution of Large Language Models (LLMs) have significantly impacted the field of natural language processing and beyond. Seminal works like BERT~\cite{devlin-etal-2019-bert} and GPT-2~\cite{radford2019language} laid the foundation for understanding context and semantics in textual language. The advent of GPT-3~\cite{NEURIPS2020_1457c0d6} demonstrated remarkable abilities in generating human-like text, paving the way for more advanced models like GPT-4~\cite{openai2023gpt4} and open-sourced Llama-2~\cite{touvron2023llama}. These models, with vast number of parameters and  pre-trained on gaint corpus, have shown exceptional skills in understanding and solving various tasks in a zero- or few-shot manner~\cite{sun2023survey}.

\subsection{Fine-tuning vs. In-Context Learning}
Fine-tuning (FT) has been a predominant approach for adapting Pre-trained Language Models (PLMs) to specific tasks, e.g., dialogue system~\cite{xu2022cosplay}. This process involves additional training of a PLM on a smaller, task-specific dataset. This technique has been proven effective in tailoring models like BERT and GPT-2 for specialized applications, from classification tasks~\cite{devlin-etal-2019-bert,liu2019roberta} to generation tasks~\cite{yang-etal-2021-contrastive-representation,gehrmann-etal-2019-generating}. 
In the era of LLMs, in-context learning (ICL)~\cite{kossen2023incontext,han-etal-2023-understanding} has emerged as a novel paradigm for distilling knowledge from powerful models, particularly highlighted in LLMs like GPT-3~\cite{NEURIPS2020_1457c0d6} and Llama-2~\cite{touvron2023llama}. ICL leverages the models' intrinsic capabilities to understand and generate responses, most of which are learned during unsupervised pre-training~\cite{zhou2023lima,gudibande2023false}, based on given contexts enclosed in the prompt, without the need for explicit task-specific training.
For instance, \citet{NEURIPS2020_1457c0d6} demonstrate the effectiveness of ICL in a wide range of tasks, by merely providing a few examples within the prompt.

The debate between FT and ICL hinges on the trade-off between specialization and generalization.
While FT offers more tailored and often higher-performing models for specific tasks, it can lead to a loss of the model's generalization abilities, as discussed by \citet{chen-etal-2020-recall}. ICL, on the other hand, maintains the model's broad applicability but may exhibit suboptimal performance in specific tasks, as observed by \citet{shi2023specialist}.
Recently, \citet{mosbach-etal-2023-shot} discovered that both few-shot FT and ICL can achieve a similar generalization on out-of-domain test sets. In contrast, our work focuses on larger-set FT instead of few-shot FT.
\citet{wang2023twostage} identified that format specialization is a critical factor contributing to the diminished ICL abilities in fine-tuned LLMs. In our experiment, we observed similar phenomena, particularly when models fine-tuned on classification tasks were evaluated on generation tasks. However, in other scenarios, the impact of format specialization appeared to be less pronounced.
\citet{anil2022exploring} found that incorporating several in-context examples during FT is helpful for length generalization for text. 
We further expand this idea and find that this approach is also indeed valuable in preserving or even enhancing the fine-tuned models' generalization ability.

\section{Evaluation Design}
This study delves into the effects of task-specific fine-tuning on the generalization ability of LLMs.
We aim to uncover whether LLMs, once fine-tuned on a dataset of a particular language task, can still perform well on i) (data-level) in-domain and out-of-domain test sets of the same task type and ii) (task-level) different tasks.

\subsection{Evaluation Taxonomy}
\label{sec:taxonomy}

To comprehensively assess the performance of fine-tuned LLMs across various tasks and datasets (Table~\ref{tab:datasets}), our study encloses three distinct settings, characterized by increasing levels of generality:

\begin{enumerate}
    \setlength{\itemsep}{0pt}
    \setlength{\parskip}{0pt}
    \item \textbf{Same Task, In-domain Datasets:} Given the same fine-tuning/test task, such as summary generation, we assess the fine-tuned LLMs using datasets that are aligned with their training data (in-domain), e.g., models fine-tuned on XSum and tested on XLSum (denoted as XSum $\rightarrow$ XLSum).
    \label{design:setting1}
    \item \textbf{Same Task, Out-of-domain Datasets:} 
    Despite being evaluated on the same task, this setting focuses on the out-of-domain generalization by testing the fine-tuned LLMs on datasets with distinct features compared to the training set, e.g., XSum $\rightarrow$ PeerRead.
    \label{design:setting2}
    \item \textbf{Different Tasks:} 
    In this setting, we examine the capability of LLMs, fine-tuned on one task type, to adapt across different task types, thereby evaluating the LLMs' cross-task generalization, e.g., XSum $\rightarrow$ Socialqa (summary to question generation) and XSum $\rightarrow$ Amazon (generation to classification).
    \label{design:setting3}
\end{enumerate}

Through such varied settings (in-domain vs. out-of-domain data and same vs. different tasks), we can clearly understand the generalization ability of the fine-tuned LLMs and explore the boundaries of their applicability across different data distributions and task types. 

\begin{table}[t]
    \small
    \centering
    \resizebox{1\hsize}{!}{
    \begin{tabular}{cccc}
    \toprule
    \textbf{Task} & \textbf{Train} & \makecell{\textbf{In-domain} \\ \textbf{Test}} & \makecell{\textbf{Out-of-domain} \\ \textbf{Test}} \\
    \midrule
    \makecell{Summary \\ Generation} & XSum & \makecell{XSum \\ XLSum} &\makecell{PeerRead \\ CNN/DailyMail} \\
    \midrule
    \makecell{Question \\ Generation} & Socialqa & Socialqa & \makecell{Tweetqa \\ Sciqa} \\
    \midrule
    \makecell{Sentiment \\ Classification} & Amazon  & \makecell{Amazon\\ AmazonFood} & \makecell{SST2 \\ Yelp} \\
    \midrule
    \makecell{Paraphrase\\ Detection} & Paws & Paws & \makecell{QQP \\ STS-B} \\
    \midrule
    \makecell{Natural Langu-\\age Inference} & MNLI & \makecell{MNLI-1 \\ MNLI-2} & \makecell{RTE \\ GPTNLI} \\
    \bottomrule
    \end{tabular}
    }
    \caption{Summary of tasks \& datasets}
    \label{tab:datasets}
    \vspace{-1em}
\end{table}

\subsection{Evaluation Benchmarks}

To achieve a comprehensive and reliable evaluation, we carefully select the evaluation benchmarks, ensuring the validity and generality of the experimental findings.
In detail, our study encompasses five widely used language tasks: summarization and question generation, sentiment classification, natural language inference, and paraphrase detection. These tasks can be broadly categorized into \textit{generation} and \textit{classification}, i.e., the first two focus on text generation, and the last three on classification.

As shown in Table~\ref{tab:datasets}, for each task, we select three or four datasets, where one is used for fine-tuning the LLMs, and the others serve as the test sets. We endeavor to ensure that the test datasets are within the same domain or span different domains as the training set. This approach can evaluate the LLMs' generalization in familiar contexts and to new domains (in- and out-of-domain).
We briefly introduce the tasks and datasets below. The full descriptions can be found in Appx.~\ref{appx:dataset}.

\paragraph{Summary Generation}
This task aims to generate a summary based on the given article.
We select XSum~\citep{Narayan2018xsum}, XLSum~\citep{hasan-etal-2021-xl}, PeerRead~\citep{kang18naacl}, and CNN/DailyMail~\citep{hermann2015CNNDailyMail}. XSum is used as the training set. The test set of XSum itself and XLSum serve as the in-domain test sets. The others are regarded as out-of-domain datasets.

\paragraph{Question Generation}
Given a paragraph and an answer, question generation infers the corresponding question.
We select Socialqa~\citep{sap2019socialiqa}, Tweetqa~\citep{xiong2019tweetqa}, and Sciqa~\citep{Welbl2017CrowdsourcingMC}.
We choose Socialqa as the training set and its test set for in-domain testing. Tweetqa and Sciqa are used as the out-of-domain test sets.

\paragraph{Sentiment Classification}
Sentiment classification identifies the positive/negative emotions expressed in the text.
This evaluation involves Amazon review~\citep{keung-etal-2020-multilingual}, AmazonFood review~\citep{amazonfood}, SST2~\citep{wang2018glue}, and Yelp\footnote{\url{https://huggingface.co/datasets/yelp_review_full}}.
Note that the target for Yelp is a rating score ranging from 1 to 5, while the label for other datasets is positive or negative. We convert the ratings in Yelp to binary labels with negative below 3.5 and positive above 3.5.
Amazon is used to fine-tune LLMs; Amazon and Amazonfood act as the in-domain test dataset, while SST2 and Yelp are the out-of-domain sets.

\paragraph{Paraphrase Detection}
This task involves classifying if two given text segments using different wordings express the same meaning.
We select Paws~\citep{zhang-etal-2019-paws}, QQP, and STS-B~\citep{wang-etal-2018-glue}.
Since STS-B is labeled using 1 to 5 similarity scores, we perform a similar processing step as YELP, i.e., if the rating is above 3.5, the two texts are paraphrased.
The Paws and itself are used as the training and in-domain test set. The other datasets are the out-of-domain datasets.

\paragraph{Natural Language Inference}
Given a pair of text segments, typically referred to as premise and hypothesis, this task determines the relationship between them, i.e., if the hypothesis is entailment, contradiction, or neutral based on the premise information.
We use MNLI~\cite{williams-etal-2018-broad} as the training set, MNLI matched (MNLI-1) and MNLI mismatched (MNLI-2) as the in-domain test sets, RTE~\cite{wang-etal-2018-glue} and GPTNLI\footnote{\url{https://huggingface.co/datasets/pietrolesci/gpt3_nli}} as the out-of-domain test sets.

\subsection{Experimental Setup}

\paragraph{Models \& Metrics}
We conduct all experiments using the open-sourced Llama-2-7b~\cite{touvron2023llama} due to its popularity in the NLP community. In the evaluation, we use the Rouge-L\footnote{\url{https://huggingface.co/spaces/evaluate-metric/rouge}} metric for generation tasks and the accuracy for classification tasks.

\paragraph{Training Details}
For each task-specific training set, we fine-tune the Llama-2 models using subsets of varying sizes: 2,000, 4,000, and 6,000 samples, which enables us to analyze how different training sizes impact the model's performance.

To standardize our training process, we treat classification tasks as text generation. Specifically, we use the language modeling head to predict labels in the text form during training, as suggested in previous works~\cite{schick-schutze-2021-just,liu-etal-2022-p}.
During the evaluation phase, we only choose the probabilities associated with these predefined labels as the models' predictions and then select the output with the highest probability as the predicted label.
Details about the labels used for classification tasks and prompt formats can be found in Appx.~\ref{appx:label} and Appx.~\ref{appx:promt}, respectively.

The models are fine-tuned with 2 epochs. We employ the AdamW~\cite{loshchilov2018decoupled} optimizer with a learning rate of 0.002. The generation length is set to 60 for generation tasks and 5 for classification tasks.

\begin{figure*}[tp]
  \centering
  \setlength{\abovecaptionskip}{2pt}
  \includegraphics[width=0.85\hsize]{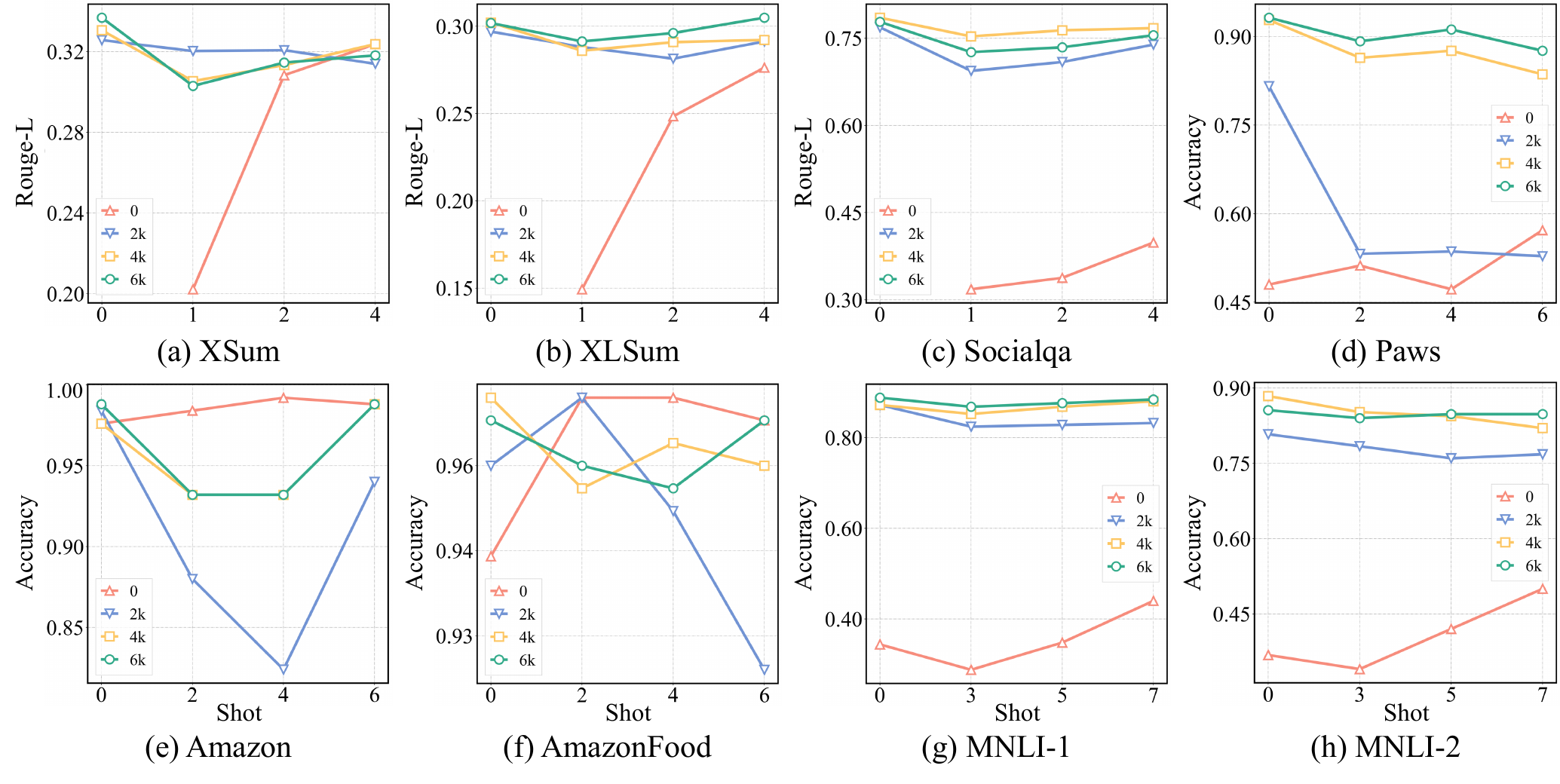}
  \caption{\textbf{In-domain} dataset testing performance comparisons of baseline Llama-2 ($0$ training samples, \textcolor{myred}{orange line}) and its fine-tuned variants ($2K, 4K, 6K$ training samples). Shot denotes the number of in-context examples. The caption for each subfigure refers to the test set. The corresponding training set can be found in Table~\ref{tab:datasets}.  The 0-shot results of the baseline Llama model in (a), (b) and (c) are not presented since in scenarios where in-context examples are absent (0-shot), baseline models generally struggle to execute the tasks effectively, even when the prompt explicitly outlines the task requirements.}
  \label{fig:in-domain}
  \vspace{-1em}
\end{figure*}

\paragraph{Testing Details}
Our primary aim is to evaluate the generalization ability of the fine-tuned LLMs.
For this, we employ distinct approaches based on the testing datasets and task types, as in Sec.~\ref{sec:taxonomy}.
When evaluating the models on datasets that match the task type of the fine-tuning set (Setting~\ref{design:setting1} and~\ref{design:setting2}), we adopt two strategies: a 0-shot prompting approach with no in-context examples (normal testing after training), and in-context learning (ICL), where a set of in-context examples are provided.
On the other hand, in scenarios where the testing task type of the test sets differs from that of the training set (Setting~\ref{design:setting3}), the usage of ICL becomes necessary. This technique is essential to inform the model about the nature of the task to be performed. Notably, for classification tasks, we still show 0-shot inference performance, since only the label probabilities are considered.

Specifically, for generation tasks, the models are evaluated with 1, 2, or 4 in-context examples. For binary classification tasks, including sentiment analysis and paraphrase detection, we report results using 2, 4, and 6 in-context examples. Regarding Natural Language Inference (NLI), which contains three label categories, we showcase results with 3, 5, and 7 in-context examples. Among all these evaluation variations, we ensure that every category is presented at least once in the in-context examples.

\section{Results and Findings}

\subsection{Same Task, In-domain Datasets}
In Figure~\ref{fig:in-domain}, we first present the results of fine-tuned Llama-2 models on in-domain testing datasets of the same fine-tune/test task type (Setting~\ref{design:setting1}). The following key findings can be drawn:

\paragraph{Fine-tuned models without in-context learning (ICL) can generally perform better than baseline Llama-2 using ICL.}
The fine-tuned models, trained with varying sample sizes (2K, 4K, 6K), exhibit superior 0-shot (w/o ICL) performance compared to the original baseline Llama2 using ICL across most datasets, notably XLSum, Socialqa, MNLI-1, MNLI-2, and Paws; see Figure~\ref{fig:in-domain} (b) (c) (d) (g) and (h).
The only exception is observed in the sentiment classification task (Amazon and AmazonFood), where fine-tuned models slightly underperform compared to baseline Llama-2 using ICL.
This could be attributed to Llama-2's inherent expertise in sentiment analysis, as indicated by its high 0-shot performance. In such cases, additional fine-tuning might not significantly enhance or could potentially impair performance, possibly due to overfitting or conflicting training data.
\paragraph{Fine-tuned LLMs often perform worse using in-context learning than the zero-shot setting.}
From Figure~\ref{fig:in-domain}, compared to baseline Llama-2, the fine-tuned models benefit little from the ICL.
This trend suggests that while fine-tuning enhances a model's 0-shot in-domain generalization, the additional in-context examples during inference are not always necessary and helpful. The drop in performance might be due to the model becoming more specialized after fine-tuning, reducing its adaptability to new contexts. The performance correlation with the number of in-context shots remains unclear for the fine-tuned models.
In comparison, baseline Llama-2 shows an overall positive tendency in performance with increased in-context shots. This improvement trend suggests that LLMs without task-specific fine-tuning are more effective in leveraging the in-context examples to understand the specialized task.
These observations can indicate that once LLMs have been fine-tuned on sufficient data, their ability to benefit from ICL diminishes.

\paragraph{Fine-tuning with more samples may not consistently improve performance on the test set.}
Generally, the models fine-tuned with 4K or 6K samples outperform the 2K models on most test sets.
However, the degree of improvement is not consistent. On XSum (Figure~\ref{fig:in-domain}~(a)), for instance, fine-tuned models demonstrate only a slight performance increase with larger training sets, changing from 2K to 6K training examples.
In contrast, on Paws (Figure~\ref{fig:in-domain}~(d)), models show marked performance gains as the number of training samples increased from 2K to 4K, with accuracy jumping from 81.6\% to 93.2\%.
We also find that increasing the training size from 4K to 6K brings subtle or even negative impacts.
These findings suggest that the relationship between the volume of fine-tuning data and in-domain test performance is task-dependent and not straightforward.

\begin{figure*}[tp]
  \centering
  \setlength{\abovecaptionskip}{2pt}
  \includegraphics[width=\hsize]{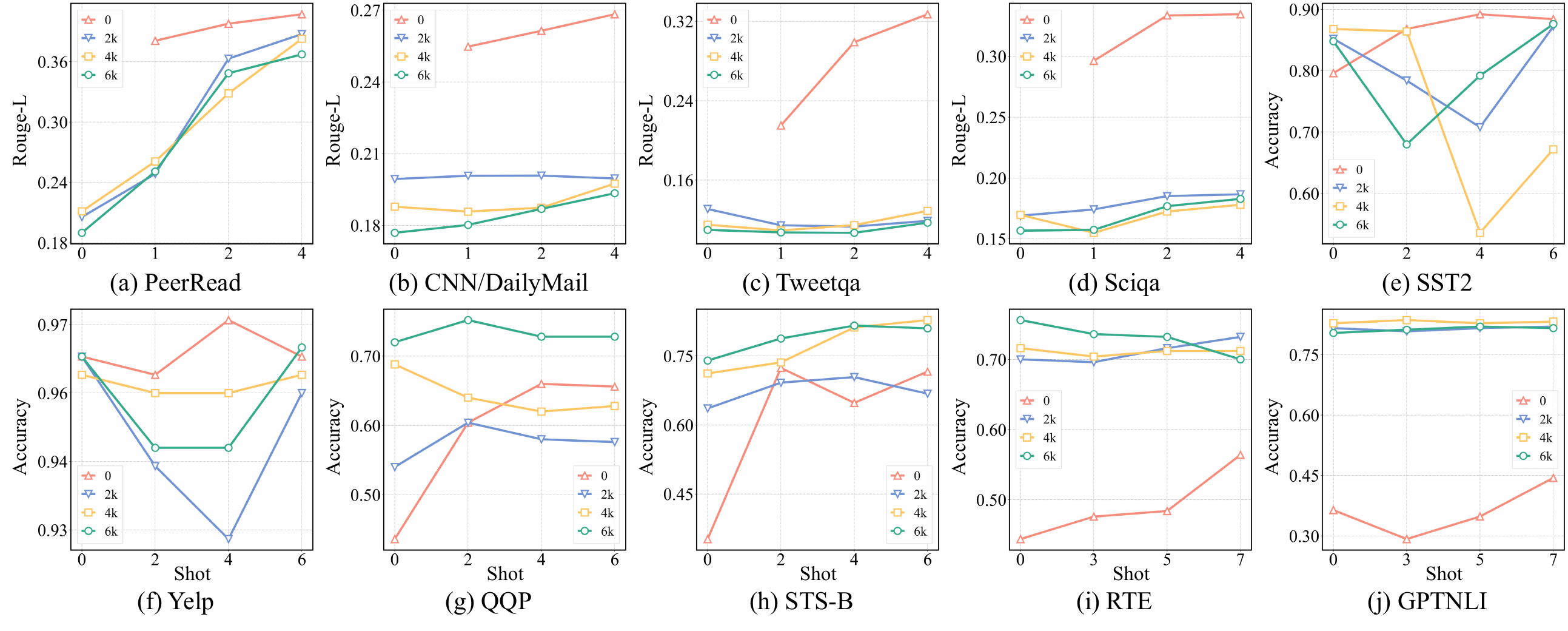}
  \caption{\textbf{Out-of-domain} dataset testing performance comparisons of baseline Llama-2 ($0$ training samples, \textcolor{myred}{orange line}) and its fine-tuned variants ($2K, 4K, 6K$ training samples).}
  \label{fig:out-domain}
  \vspace{-1em}
\end{figure*}
  
\subsection{Same Task, Out-of-domain Datasets}
\label{sec:out-of-domain}
We then show the results of the fine-tuned Llama-2 on out-of-domain testing datasets of the same task type (Setting~\ref{design:setting2}). We derive the following observations based on Figure~\ref{fig:out-domain}:

\paragraph{Fine-tuned models underperform compared to the baseline model on generation tasks, yet outperform on classification tasks.}
For the out-of-domain testing results, a clear distinction emerges based on the task categories, i.e., between generation and classification tasks.
In generation testing datasets ((a)~PeerRead, (b)~CNN/DailyMail, (c)~Tweetqa, and (d)~Sciqa in Figure~\ref{fig:out-domain}), fine-tuned models perform worse than baseline models, and this gap persists regardless of the number of in-context examples provided. Notably, there lacks a performance growth with more training examples in datasets like CNN/DailyMail, Tweetqa and Sciqa, suggesting that the generalization ability of fine-tuned LLMs is impaired.

On the other hand, in the context of sentiment classification tasks, the best performance of the fine-tuned models is on par with the baseline Llama-2, which means the fine-tuned LLMs' sentiment knowledge is unlikely to be lost; see Figure~\ref{fig:out-domain}~(e) and (f). 
Another notable observation is that the fine-tuned models exhibit larger variability using different ICL shots. This indicates that the performance is more sensitive to the in-context examples for the sentiment classification task.
For other classification tasks (i.e., paraphrase detection and natural language inference), fine-tuned models consistently outperform the baseline Llama-2, as shown in Figure~\ref{fig:out-domain} (g)-(j).

The diverging effects of fine-tuning on generation and classification tasks for out-of-domain testing may originate from the difference in task output space constraints.
The output space of classification tasks is inherently predefined and limited, enabling fine-tuned LLMs to apply their inherited and adapted knowledge relatively easily to new domains.
In contrast, the output space of out-of-domain generation datasets largely deviates from that of the training set. Despite being given a few in-context examples, fine-tuned models may still find it challenging to reason about the expansive range of possible outputs in new domains.

\begin{figure*}[h!]
  \centering
  \setlength{\abovecaptionskip}{2pt}
  \includegraphics[width=\hsize]{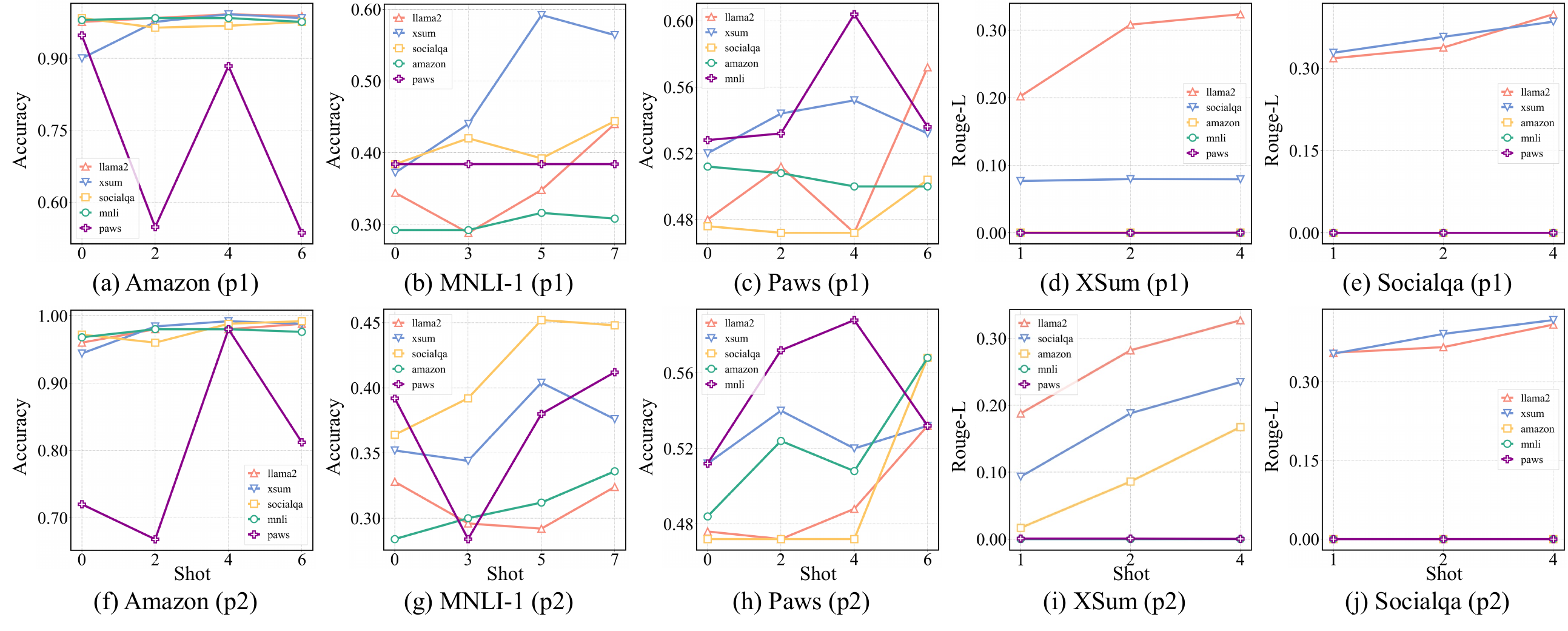}
  \caption{\textbf{Cross-task} performance comparisons of baseline Llama-2 ($0$ training samples, \textcolor{myred}{orange line}) and models fine-tuned on other tasks. The caption for each subfigure refers to the test set. The legends denote the training data. The first row is the results using the Prompt-1 (p1) and the second row is the results using the Prompt-2 (p2) format. The detailed prompt formats can be found in Appx.~\ref{appx:promt}.}
  \label{fig:cross-domain}
  \vspace{-1em}
\end{figure*}

\subsection{Different Tasks}
\label{sec:difftasks}
Last, we test whether the generalization ability of fine-tuned models is preserved on cross-task testing datasets (Setting~\ref{design:setting3}).
The evaluation for each testing set is performed using models fine-tuned on 2K training samples from other tasks.
Due to space limitations, our analysis is confined to five test sets: XSum, Socialqa, Amazon, Paws, and MNLI-1.
The following findings are mainly based on the first row of Figure~\ref{fig:cross-domain}:

\paragraph{The generalization through fine-tuning exhibits significant variability and highly depends on the training data.}
When assessing performance on classification tasks, i.e., Amazon, MNLI, and Paws, it becomes evident that the model's fine-tuning source greatly affects its efficacy. Fine-tuning on a dataset like Amazon negatively impacts performance on the MNLI-1 test set, whereas fine-tuning on XSum significantly boosts it; see Figure~\ref{fig:cross-domain} (b).
In parallel, for generation tasks, training on Socialqa hurt the performance on XSum while training on XSum has little impact on Socialqa; see Figure~\ref{fig:cross-domain} (d) and (e).
These intricate patterns suggest that the effectiveness of fine-tuning is not easily predictable and likely intertwines with dataset characteristics, fine-tuning procedures, etc.

\paragraph{Models fine-tuned on classification tasks fail to generalize to generation tasks.}
From Figure~\ref{fig:cross-domain} (d) and (e), fine-tuning the models on classification data leads to almost zero Rouge-L scores for generation tasks. Upon examining the outputs, it becomes evident that these models predominantly generate classification labels rather than coherent text, a manifestation of output space specialization, which aligns with the findings of \citet{wang2023twostage}.
Two potential reasons exist. Firstly, the model's output space may be constrained to the category labels seen during fine-tuning, inhibiting its ability to generate other tokens. The second may be induced by the prompt format, as listed as Prompt-1 in Table~\ref{tab:prompt}. The prompts for different tasks have the same start "\#\#\#", which could confuse the model and cause it to misinterpret inputs from other tasks as if they belong to its training task.
To further investigate the influence of prompt format, we introduced a distinct set of prompts that avoid uniform starting sequences (Prompt-2 in Table~\ref{tab:prompt}). The corresponding results are shown in the second row of Figure~\ref{fig:cross-domain}.
Comparing the first and second rows, we find that the cross-task evaluation on classification tasks is more sensitive to the prompt format than the generation task evaluation.
Moreover, from Figure~\ref{fig:cross-domain}~(i), the model fine-tuned on Amazon starts to work on XSum with such new prompts. Yet, for Socialqa (subfigure~(j)), the fine-tuning on Amazon still fails to succeed.
Hence, the prompt format is crucial for cross-task generalization, but identifying a universally effective format remains to be explored.

\section{Fine-tuning with In-context Learning on Generation Tasks Helps Improve the Generalization Ability of LLMs}

In Sec.~\ref{sec:out-of-domain}, our analysis reveals that LLMs fine-tuned on classification tasks exhibit robust generalization capabilities under out-of-domain scenarios.
However, for generation tasks, the fine-tuned models consistently fall short compared to baseline.

In this section, we aim to show that Fine-Tuning with In-Context Learning (FTICL) can help improve LLMs' out-of-domain generalization for generation tasks.
FTICL leverages the strengths of both fine-tuning and ICL. Specifically, instead of directly forwarding the original input into LLMs during fine-tuning, FTICL prepends in-context examples to the input and forwards the LLMs with this newly constructed input.
Note that this strategy has also been adopted to mitigate other fine-tuning issues. For example, \citet{anil2022exploring} find that FTICL leads to a substantial improvement in terms of input length generalization.

\begin{figure*}[t]
\setlength{\abovecaptionskip}{2pt}
  \centering
  \includegraphics[width=0.85\hsize]{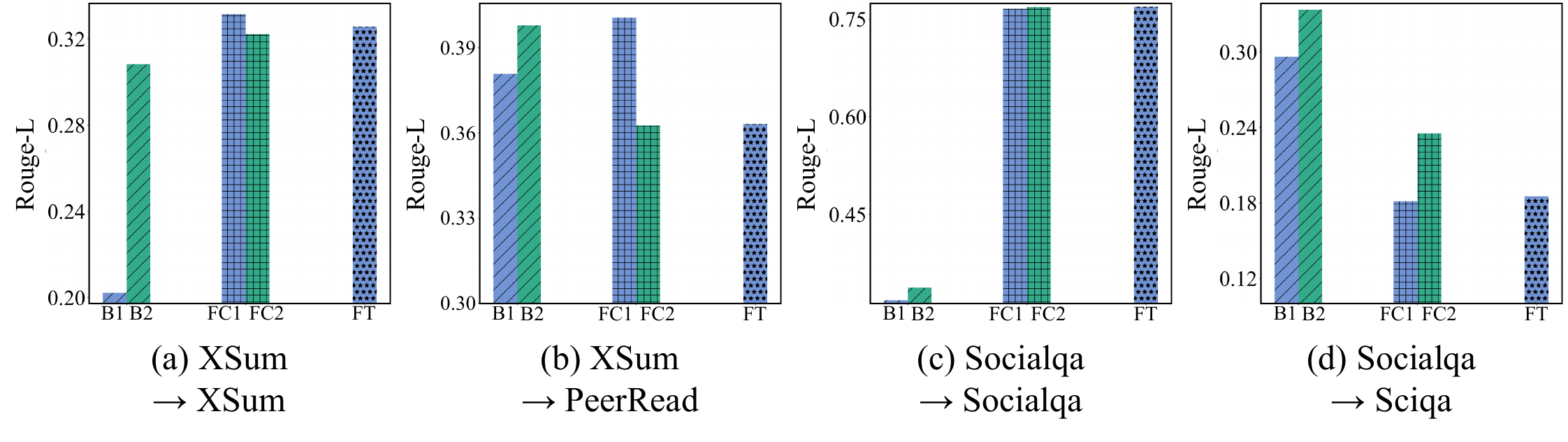}
  \caption{\textbf{Same fine-tuning/test task type evaluation of FTICL with generation tasks.} B$n$ represents the baseline Llama-2 model with $n$ in-context examples during inference. FC$n$ denotes the FTICL models fine-tuned with $n$ in-context examples. FT is the vanilla fine-tuned model without in-context learning. For FC$n$ and FT, we fine-tune with 2,000 samples, perform both 0-shot and few-shot evaluations, and report the results with the best performance.}
  \label{fig:ptune-same-task}
\end{figure*}

\begin{figure*}[h]
  \centering
  \setlength{\abovecaptionskip}{2pt}
  \includegraphics[width=\hsize]{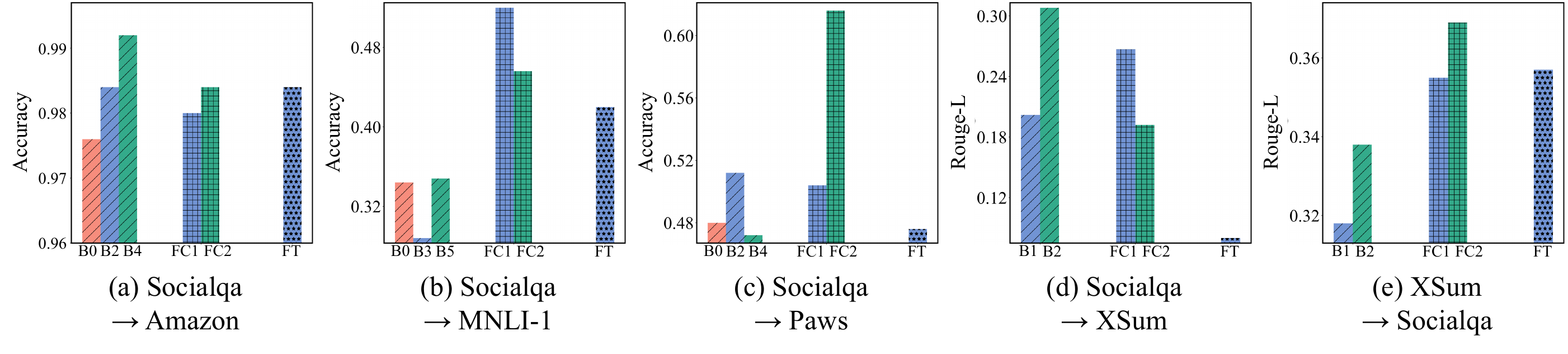}
  \caption{\textbf{Cross-task performance of FTICL with generation tasks.} For the classification task evaluation, we also report the 0-shot performance (B0) for the baseline Llama-2.
  }
  \label{fig:ptune-cross}
  \vspace{-1em}
\end{figure*}

In our experiment, we fine-tune the FTICL models using 2,000 samples. For each generation task, we train two models with 1 or 2 in-context examples prepended in input, respectively. The results are presented in Figure~\ref{fig:ptune-same-task} and Figure~\ref{fig:ptune-cross}.
The following findings regarding generalization are organized in terms of the same or different fine-tuning/test tasks.

\subsection{Same Task}
We investigate the effects of FTICL across datasets sharing the same fine-tuning/test task type.
We fine-tune models on the training sets of the XSum and Socialqa datasets, respectively. These models are then evaluated both on their respective test sets and on distinct out-of-domain test sets.
The main results of these evaluations are presented in Figure~\ref{fig:ptune-same-task}. Due to space limitations, we only report the typical results on XSum, PeerRead, Socialqa, and Sciqa. 
From subfigures (a) and (c), we can see that FTICL models preserve or even slightly enhance the performance on the corresponding test sets of the training set compared with vanilla fine-tuning.

When we extend our analysis to out-of-domain test sets of the same fine-tuning/test task type, we find \textbf{models fine-tuned using FTICL achieve a better out-of-domain generalization performance than the vanilla fine-tuned models}. The FTICL models can sometimes even surpass the baseline model without fine-tuning.
For instance, as shown in Figure~\ref{fig:ptune-same-task}~(b), the FTICL model with one in-context example during fine-tuning (FC1) showcases a remarkable gain over the vanilla fine-tuned model (FT) on PeerRead, surpassing even the baseline Llama-2 model using in-context learning (B1 and B2).
For Sciqa, although FC2 lags behind B1 and B2, it still performs better than FT; see Figure~\ref{fig:ptune-same-task}~(d).
These observations reveal that FTICL may mitigate catastrophic forgetting for generation tasks by allowing the model to retain its learned capabilities more effectively than the vanilla fine-tuning method.

\subsection{Different Tasks}
Besides enhanced out-of-domain generalization, we also demonstrate that models utilizing FTICL exhibit superior cross-task generalization capabilities. The results are shown in Figure~\ref{fig:ptune-cross}.

\paragraph{FTICL trained on generation tasks achieves a better cross-task generalization than vanilla fine-tuning.}
We can observe FTICL models fine-tuned on Socialqa achieve at least comparable performance on Amazon, shown in Figure~\ref{fig:ptune-cross} (a), and exhibit superior results over both the baseline Llama-2 and the vanilla fine-tuned models when evaluated on MNLI and Paws; see Figure~\ref{fig:ptune-cross} (b) and (c).
We can also see FC1 in subfigure (d) and FC1, FC2 in subfigure (e) outperform the vanilla fine-tuned (FT) models and can be even on par with or surpass the baseline Llama-2 for the cross summary and question generation task evaluations. The results show that fine-tuning with ICL is better than directly fine-tuning on generation tasks.

\subsection{Potential Reason}

We provide one hypothesis that could drive the success of FTICL in enhancing LLMs' generalization:
\textbf{FTICL tends to deviate less from the original LLM than vanilla fine-tuning.}
In other words, FTICL models preserve more general knowledge inherent in LLMs.
To support this, we calculate the average parameter weight difference between the fine-tuned models (FTICL and FT) and the original Llama-2.
Experimental results show consistency with our hypothesis: on Socialqa, FTICL~($7.95e-05$) vs. FT~($8.54e-05$); on XSum, FTICL~($8.03e-05$) vs. FT~($1.0e-04$).
The potential reason is that the provided in-context examples encourage the LLM to leverage its existing knowledge to solve new tasks.

\subsection{FTICL on Classificaion Tasks}
\begin{figure*}[h]
\setlength{\abovecaptionskip}{2pt}
  \centering
  \includegraphics[width=\hsize]{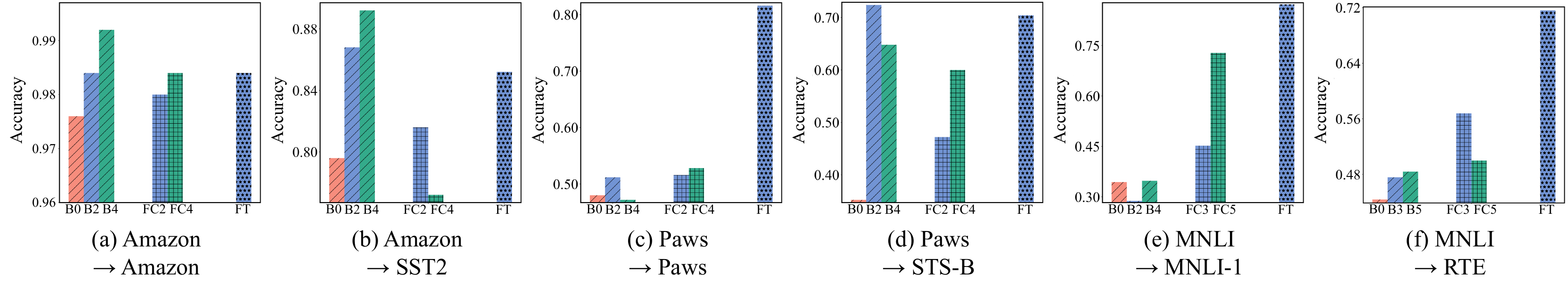}
  \caption{\textbf{Same fine-tuning/test task type evaluation of FTICL models fine-tuned on classification tasks.}}
  \label{fig:ptune-class}
\end{figure*}

\begin{figure*}[h]
\setlength{\abovecaptionskip}{2pt}
  \centering
  \includegraphics[width=\hsize]{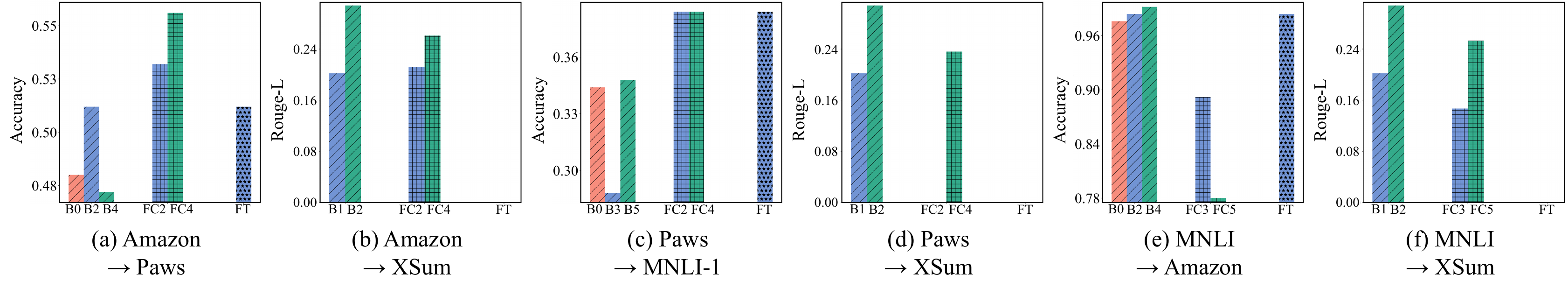}
  \caption{\textbf{Cross-task performance for FTICL models fine-tuned on classification tasks.}}
  \label{fig:ptune-class-cross}
  \vspace{-1em}
\end{figure*}

We also analyse the performance of FTICL on classification tasks  and find that this strategy \textbf{does not} help improve the generalization ability of LLMs as shown in Figure~\ref{fig:ptune-class} and Figure~\ref{fig:ptune-class-cross}.

For the classification tasks, as shown in Figure~\ref{fig:ptune-class} (a) (c), and (e), when evaluated on the in-domain test set of the corresponding training set, fine-tuning with in-context learning (FTICL) models generally perform worse than vanilla fine-tuned models. For the out-of-domain test sets, we find FTICL models can generally outperform the baseline Llama but lag behind vanilla FT models; see Figure~\ref{fig:ptune-class} (b) (d) (f). The reason for this phenomenon may be that classification tasks are more sensitive to the in-context examples. We also hypothesize that the optimization process plays a pivotal role. Specifically, for classification tasks, we observed that the final loss of models fine-tuned with in-context learning (FTICL) tends to be higher compared to vanilla fine-tuning (FT), with challenges in further reducing the loss. This phenomenon might stem from the model's tendency to be lazy. Specifically, for generation tasks (e.g., summary), the model must learn to leverage contextually relevant information (e.g., the article to be summarized) to generate appropriate outputs (e.g., target summary). However, for classification tasks, in-context examples could inadvertently act as distractors since all the labels are provided in the in-context examples. It may lead the model to copy labels directly from in-context examples rather than leveraging the corresponding relevant information. This could also explain the difficulty in loss reduction. A better optimizer could possibly solve this problem.
The above observations suggest that for classification tasks, it is better to adopt the vanilla fine-tuning approach instead of FTICL.

Lastly, Figure~\ref{fig:ptune-class-cross} shows the cross-task performance of FTICL, where the models are fine-tuned with classification tasks. In Figure~\ref{fig:ptune-class-cross} (a) (c), and (e), for cross-task classification tasks, we find the cross-task transfer effects are not clear. For example, training on Amazon has a positive effect on Paws, while training on MNLI has a negative effect on Amazon. 
Meanwhile, as shown in Figure~\ref{fig:ptune-class-cross} (b) (d), and (f), we can see that FTICL models for generation task evaluation outperform vanilla fine-tuned models, which means FTICL can help alleviate the output space specialization issue mentioned in Sec.~\ref{sec:difftasks}.

\section{Conclusion}
This study comprehensively investigates the effects of fine-tuning on the LLMs' generalization ability.
We conduct systematic experiments by evaluating the fine-tuned LLMs across various training data and language tasks.
Experimental results indicate that dissimilar generalization ability after fine-tuning arises from the nature of generation or classification tasks.
Further, we show that fine-tuning with in-context learning can enhance the generalization capability for generation tasks.
We hope our findings can inspire future advancements in understanding and effectively utilizing LLMs to solve new tasks.

\section*{Limitations}
In this work, we study the LLMs' generalization ability between fine-tuned variants and their original counterparts and explore the effective fine-tuning strategies.
This work has two limitations:
i)~The underlying reasons for the difference between models fine-tuned on classification tasks and generation tasks remain under-examined.
ii)~The intricate operational mechanisms behind the fine-tuning with in-context learning method have yet to be exhaustively understood.
We leave these more in-depth analyses as our future work.


\bibliography{anthology,custom}

\clearpage
\appendix

\label{sec:appendix}

\section{Dataset Information}
\label{appx:dataset}
We provide more dataset information (including the data sources and examples) in Table~\ref{tab:data-1} and Table~\ref{tab:data-2}. It should be noted that for the summary generation task, although both XSum~\citep{Narayan2018xsum} and CNN/DailyMail~\citep{hermann2015CNNDailyMail} are from new articles, they are not in the same domain.
There are two reasons. First, XSum is from BBC News, and CNN/DailyMail is from CNN and DailyMail news. There exist style differences for different news. Another more important reason is that XSum is a one-sentence summary dataset, while CNN/DailyMail is a multiple-sentence summary dataset; see Table~\ref{tab:data-1}.

\begin{table*}[t]
    \small 
    \centering
    \resizebox{1\hsize}{!}{
    \begin{tabular}{p{0.12\hsize} p{0.12\hsize} p{0.12\hsize} p{0.5\hsize}}
    \toprule
    Task & Dataset & Source & Example \\
    \midrule 
    \multirow{22}{\hsize}{Summary Generation} & \multirow{5}{\hsize}{XSum \citep{Narayan2018xsum}} & \multirow{5}{\hsize}{BBC news} & \textbf{Input:} Electrician Carl Holdsworth has set up holographic video footage of Mr and Mrs Claus behind the windows of his Chaddesden house... \\
                                              &                            &                                & \textbf{Output:} Festive revellers have travelled for miles to see Father Christmas and his wife apparently living in a Derby home. \\
                                              \cmidrule{2-4}
                                              & \multirow{6}{\hsize}{XLSum \citep{hasan-etal-2021-xl}} & \multirow{6}{\hsize}{BBC news} & \textbf{Input:} Jack McLinden, who has multiple health conditions, experienced joining his heroes on the pitch before their game against Newcastle United on Monday... \\
                                              &                             &                                & \textbf{Output:} A 14-year-old Everton fan has made history by becoming football's first ``remote'' match-day mascot - with the aid of a robot. \\
                                              \cmidrule{2-4}
                                              & \multirow{4}{\hsize}{PeerRead \citep{kang18naacl}} & \multirow{4}{\hsize}{Scientific peer reviews} & \textbf{Input:} We explore techniques to maximize the effectiveness of discourse information in the task of authorship attribution... \\
                                              &                                &                                               & \textbf{Output:} Leveraging Discourse Information Effectively for Authorship Attribution. \\
                                              \cmidrule{2-4}
                                              & \multirow{6}{\hsize}{CNN/DailyMail \citep{hermann2015CNNDailyMail}} & \multirow{6}{\hsize}{CNN news and DailyMail} & \textbf{Input:} Chris Brown sat alone in court for 35 minutes on Friday while his lawyer talked with the judge and prosecutor behind closed doors in his probation violation case... \\
                                              &                                     &                                              & \textbf{Output:} Judge orders Brown to come back to court on June 10. Prosecutors accuse Brown of not finishing 180 days of community labor. \\
    \midrule
    \multirow{13}{\hsize}{Question Generation} & \multirow{4}{\hsize}{Socialqa \citep{sap2019socialiqa}} & \multirow{4}{\hsize}{Social Commonsense} & \textbf{Input:} Sydney is always respected by Kai. They were able to make up Kai's mind. \\
                                               &                                &                            & \textbf{Answer:} thank Sydney sincerely \\
                                               &                                &                            & \textbf{Output:} What will happen to Kai? \\
                                               \cmidrule{2-4}
                                               & \multirow{4}{\hsize}{Tweetga \citep{xiong2019tweetqa}} & \multirow{4}{\hsize}{Twitter} & \textbf{Input:} Getting taxis is a nightmare - local drivers confused with new street layout, translations on phone app! \\
                                               &                               &                            & \textbf{Answer:} getting taxis \\
                                               &                               &                            & \textbf{Output:} what does ben have nightmares of? \\
                                               \cmidrule{2-4}
                                               & \multirow{4}{\hsize}{Sciqa \citep{Welbl2017CrowdsourcingMC}} & \multirow{4}{\hsize}{School Science Textbooks} & \textbf{Input:} Archaea live everywhere on Earth, including extreme environments. \\
                                               &                             &                            & \textbf{Answer:} everywhere \\
                                               &                             &                            & \textbf{Output:} Where do archea live? \\
    \midrule
    \multirow{16}{\hsize}{Sentiment Classification} & \multirow{4}{\hsize}{Amazon \citep{keung-etal-2020-multilingual}} & \multirow{4}{\hsize}{Product review} & \textbf{Input:} I will not use it again. Made my dogs feel bad. One of my dogs lost hair because of this product. I had to end up washing my dogs to remove as must as possible. \\
                                                    &                              &                            & \textbf{Output:} negative \\
                                                    \cmidrule{2-4}
                                                    & \multirow{4}{\hsize}{AmazonFood \citep{amazonfood}} & \multirow{4}{\hsize}{Food review} & \textbf{Input:} Great taste, texture, flavor, and works well with any recipe! Wonderful pricing, too! Gluten free products are so expensive normally! Definitely recommend this! \\
                                                    &                                  &                            & \textbf{Output:} positive \\
                                                    \cmidrule{2-4}
                                                    & \multirow{4}{\hsize}{SST2 \citep{wang2018glue}} & \multirow{4}{\hsize}{Moview review} & \textbf{Input:} In its dry and forceful way, it delivers the same message as Jiri Menzel's Closely Watched Trains and Danis Tanovic's No Man's Land. \\
                                                    &                            &                            & \textbf{Output:} positive \\
                                                    \cmidrule{2-4}
                                                    & \multirow{6}{\hsize}{Yelp} & \multirow{6}{\hsize}{Yelp review} & \textbf{Input:} Service sucks! They're usually understaffed and we were told to wait an hour for a table to be cleaned. There was only 1 chef and after we sat we waited for our food for 20 minutes!!!! Even the appetizers took 20 minutes. Will never recommend this place to anyone again. Big big mistake. \\
                                                    &                            &                            & \textbf{Output:} negative \\
    \bottomrule
    \end{tabular}
    }
    \caption{Dataset Description \#1}
    \label{tab:data-1}
\end{table*}

\begin{table*}[t]
    \small 
    \centering
    \resizebox{1\hsize}{!}{
    \begin{tabular}{p{0.12\hsize} p{0.12\hsize} p{0.12\hsize} p{0.5\hsize}}
    \toprule
    Task & Dataset & Source & Example \\
    \midrule 
    \multirow{13}{\hsize}{Paraphrase Detection} & \multirow{7}{\hsize}{Paws \citep{wang-etal-2018-glue}} & \multirow{7}{\hsize}{Wikipedia} & \textbf{Input 1:} Windows XP Mode runs Windows XP in a virtual machine , and displays applications within separate windows on the Windows 7 desktop. \\
                                                &                            &                            & \textbf{Input 2:} Windows XP - Mode executes Windows XP on a separate machine and displays applications in virtual windows on the Windows 7 desktop. \\
                                                &                            &                            & \textbf{Output:} no \\
                                                \cmidrule{2-4}
                                                & \multirow{3}{\hsize}{QQP \citep{wang-etal-2018-glue}} & \multirow{3}{\hsize}{Quora Question} & \textbf{Input 1:} How do I develop the patience to read books? \\
                                                &                           &                            & \textbf{Input 2:} How can I develop patience and love towards reading? \\
                                                &                           &                            & \textbf{Output:} yes \\
                                                \cmidrule{2-4}
                                                & \multirow{3}{\hsize}{STS-B \citep{wang-etal-2018-glue}} & \multirow{3}{\hsize}{Misc.} & \textbf{Input 1:} US postpones missile test over N Korea tensions \\
                                                &                             &                            & \textbf{Input 2:} Embassies staying put in N. Korea despite tension \\
                                                &                             &                            & \textbf{Output:} no \\
    \midrule
    \multirow{18}{\hsize}{Natural Language Inference} & \multirow{7}{\hsize}{MNLI \cite{williams-etal-2018-broad}} & \multirow{7}{\hsize}{Misc.} & \textbf{Input 1:} i think that's great there's a few places in Houston where they're trying that out i don't know if it's the if they've done it citywide yet or not where they have the color coded uh bags and uh bins \\
                                                      &                            &                            & \textbf{Input 2:} There are a couple places in Houston where it's being tried. \\
                                                      &                            &                            & \textbf{Output:} entailment \\
                                                      \cmidrule{2-4}
                                                      & \multirow{8}{\hsize}{RTE \cite{wang-etal-2018-glue}} & \multirow{8}{\hsize}{d Wikipedia} & \textbf{Input 1:} The third, and most remote possibility, was considered to be a so-called Jamaican coalition, again based on the party colours, between the Christian Democrats, the FDP and the Greens. \\
                                                      &                           &                            & \textbf{Input 2:} It seems unlikely that there will be a coalition between Gerhard Schroeder's Social Democrats and Angela Merkel's Christian Democratic Union. \\
                                                      &                           &                            & \textbf{Output:} neutral \\
                                                      \cmidrule{2-4}
                                                      & \multirow{4}{\hsize}{GPTNLI} & \multirow{4}{\hsize}{Misc.} & \textbf{Input 1:} The Great Horned Owl has a tuft of feathers around its face that makes it look like an old man's white mustache. \\
                                                      &                              &                            & \textbf{Input 2:} The Great Horned Owl is bald. \\
                                                      &                              &                            & \textbf{Output:} contradiction \\
    \bottomrule
    \end{tabular}
    }
    \caption{Dataset Description \#2}
    \label{tab:data-2}
\end{table*}

\section{Classification Labels}
\label{appx:label}
In Table~\ref{tab:label}, we illustrate the tokens that need to be generated during training for the language classification tasks. We ensure the labels between different tasks have no intersection with each other. This can avoid the target specialization issue when evaluating a dataset in one task type using models trained on a dataset of a different task type.

\begin{table*}[t]
    \centering
    \begin{tabular}{cc}
    \toprule
    Task & Label\\
    \midrule 
    Sentiment Classification & positive, negative \\ 
    Paraphrase Detection & yes, no \\
    Natural Language Inference & entailment, contradiction, neutral \\
    \bottomrule
    \end{tabular}
    \caption{The generation labels for classification tasks.}
    \label{tab:label}
\end{table*}

\section{Prompt Format}
\label{appx:promt}

In Table~\ref{tab:prompt}, we detail the prompt formats employed for each task in our experiments. Throughout the training phase, the model is trained using Prompt-1. Note that Prompt-1 is also the default choice during the testing phase. To ensure a comprehensive evaluation, particularly in cross-task scenarios where the influence of the prompt format is a crucial consideration, we additionally report results obtained using Prompt-2. This approach allows us to assess the impact of different prompt structures on model performance.

\begin{table*}[ht]
    \centering
    \begin{tabular}{ccc}
    \toprule
    Task & Prompt-1 & Prompt-2 \\
    \midrule
    \makecell{Summary\\ Generation} & \makecell{\#\#\# Input: \{\textit{input}\} \\ \#\#\# Summary:} & \makecell{Please read the following \\ text: \{\textit{input}\} Provide a summary: } \\
    \midrule 
    \makecell{Question \\ Generation} & \makecell{\#\#\# Input: \{\textit{input}\} \\ \#\#\# Answer: \{\textit{answer}\} \\ \#\#\# Question:} & \makecell{Given the context:\{\textit{input}\} And \\ the answer: \{\textit{answer}\} Generate \\ a suitable question: } \\
    \midrule
    \makecell{Sentiment \\ Classification} & \makecell{\#\#\# Input: \{\textit{input}\} \\ \#\#\# Sentiment: } & \makecell{Analyze the sentiment of the \\ following text: \{\textit{input}\} Sentiment: } \\
    \midrule
    \makecell{Paraphrase\\ Detection} & \makecell{\#\#\# Input\_1: \{\textit{input\_1}\} \\ \#\#\# Input\_2: \{\textit{input\_2}\} \\ \#\#\# Paraphrase Classification:} & \makecell{Let's compare the two \\ sentences: Sentence\_1: \{\textit{input\_1}\} Sentence\_2: \\ \{\textit{input\_2}\} Are they paraphrasing?:} \\
    \midrule
    \makecell{Natural Langu-\\age Inference} & \makecell{\#\#\# Input\_1: \{\textit{input\_1}\} \\ \#\#\# Input\_2: \{\textit{input\_2}\} \\ \#\#\# Inference:} & \makecell{Consider the following texts:\\ Text 1: \{\textit{input\_1}\}
    Text 2: \{\textit{input\_2}\} \\
    The relation is } \\
    \bottomrule
    \end{tabular}
    \caption{Prompt formats for each task}
    \label{tab:prompt}
\end{table*}

\end{document}